\documentclass[10pt,twocolumn,letterpaper]{article}

\usepackage{iccv}
\usepackage{times}
\usepackage{graphicx}
\usepackage{amsthm}
\usepackage{amsmath}               
  {
      \theoremstyle{plain}
      \newtheorem{assumption}{Assumption}
  }
\usepackage{amssymb}
\usepackage{multirow}

\usepackage[breaklinks=true,bookmarks=false]{hyperref}

\iccvfinalcopy 

\ificcvfinal\fi

\begin{document}

\title{DeepFake Detection with Inconsistent Head Poses: Reproducibility and Analysis}

\author{Kevin Lutz\\
Naval Postgraduate School\\
1 University Circle, Monterey, CA 93943\\
{\tt\small kevin.lutz@nps.edu}
\and
Robert Bassett\\
Naval Postgraduate School\\
1 University Circle, Monterey, CA 93943\\
{\tt\small robert.bassett@nps.edu}
}

\maketitle
\ificcvfinal\thispagestyle{empty}\fi

\begin{abstract}
   Applications of deep learning to synthetic media generation allow the creation of convincing forgeries, called DeepFakes, with limited technical expertise. DeepFake detection is an increasingly active research area. In this paper, we analyze an existing DeepFake detection technique based on head pose estimation, which can be applied when fake images are generated with an autoencoder-based face swap \cite{thepaper}. Existing literature suggests that this method is an effective DeepFake detector, and its motivating principles are attractively simple. With an eye towards using these principles to develop new DeepFake detectors, we conduct a reproducibility study of the existing method. We conclude that its merits are dramatically overstated, despite its celebrated status. By investigating this discrepancy we uncover a number of important and generalizable insights related to facial landmark detection, identity-agnostic head pose estimation, and algorithmic bias in DeepFake detectors. Our results correct the current literature's perception of state of the art performance for DeepFake detection.
\end{abstract}

\section{Introduction}

Detecting manipulated media is a research priority for both public and private sector institutions. This broad interest reflects the fact that visual misinformation has the potential to cause grave damage in a number of areas, including financial and political systems. In recent years, applications of deep learning to image manipulation have led to the emergence of \emph{DeepFakes}, which lower the technological barriers required to create high-quality manipulations. The accessibility of this technology has led to a proportional response from researchers attempting to distinguish authentic from manipulated media, with the result that many promising methods for detecting DeepFakes have recently been developed. Though the term ``DeepFake'' is often used as a universal descriptor for any synthetic media generated wholly or partially with deep learning, in this paper we will focus on images which are created with an autoencoder-based face swap.

One particularly promising technique for detecting DeepFakes is based on finding head pose inconsistencies in modified images \cite{thepaper}. This method, which we henceforth refer to as the \emph{analytic}, was demonstrated by its authors to classify fake from real images with high accuracy. Moreover, its motivating principle is attractively simple, in contrast to many other methods for detecting DeepFakes which lack interpretation. These merits have led to the analytic becoming extremely popular, garnering nearly 200 citations in the two years since its publication.

In an attempt to objectively assess the utility of this analytic for DeepFake detection, we conduct an in-depth analysis of its underlying methodology, in addition to a reproducibility study. We find a stark lack of generalizability of the analytic's performance to any data other than those originally included in the paper. Moreover, we are able to \emph{explain} this lack of generalizability by identifying a number of incorrect assumptions from the original manuscript. The first of these is related to the utility of estimated head poses as a feature. We show that, surprisingly, pose estimates contain enough information to identity unique individuals. It follows that, when trained on faces which exhibit similar structure (for example, due to a shared ethnicity or gender), this analytic can exhibit algorithmic bias because of its tendency to classify images as authentic or manipulated based on the facial structure of its subject. Another incorrect assumption we identify relates to the head pose estimation itself. We show that, without additional modifications, iterative methods for pose estimation tools often fail to accurately estimate head poses because of their tendency to get stuck in a strong local minimum. The analytic demonstrates this behavior, resulting in head pose estimates which are frequently very poor. We propose a simple correction which avoids this local minimum. Although it does not remedy the performance issues of the analytic, we expect this contribution will have utility in other contexts which require head pose estimation as a computational primitive.

The rest of this paper is organized as follows. In the next subsection we outline related work, and in section \ref{sec:review} we review the analytic. Section \ref{sec:assumps} introduces and then refutes four assumptions necessary to the analytic's rationale. In section \ref{sec:reprod}, we conduct a set of numerical experiments on a variety of DeepFake data sets. We conclude with section \ref{sec:conclude}, where we summarize our contributions and their implications.


\subsection{Related Work}

One of the most common tools for generating DeepFakes is the autoencoder-based face swap \cite{DeepFaceLab, Dfaker, faceswap-gan, faceswap, fakeapp}. We will focus on this technique exclusively in this paper, though other techniques, such lip-sync or puppet master, also exist for creating DeepFakes \cite[\S 2.1]{farid}. To perform a face swap between two individuals, faces must be detected, aligned, and segmented from a collection of images of each individual. Then, an autoencoder network is adversarially trained on the aligned faces, where a single encoder network is used for both individuals but a unique decoder is used for each. After training, one can swap the face of the first individual into an image of the second as follows. First, an image of the second individual is compressed using the encoder. To perform the swap, the encoded image of the second individual is decoded using the decoder trained on the first individual. Afterwards, postprocessing is an important final step to eliminate visual artifacts where the swapped inner region and the original outer portion of the face meet \cite{Celeb-DF}.


Motivated by the convincing nature of modern DeepFakes, many methods have been proposed to detect them from authentic images. These methods can be partitioned by the features they use to detect the manipulated image. Though not specific to DeepFake detection, methods which are designed to detect image splicing, where a part of a source image is placed into a target image, can be applied for this purpose. Examples include \cite{splice1}, where an autoencoder is trained to partition a face into spliced and unmanipulated regions, and \cite{splice2}, where image metadata is leveraged to detect inconsistencies and generate a splice mask for various regions of an image. In \cite{splice3}, the authors take a different perspective by formulating a hypothesis test for each pixel, using $Z$-scores constructed for various neighborhoods of the pixel as a test statistic. In \cite{splice4}, the authors detect spliced regions using steganalysis features unique to the processing of individual cameras by comparing these features in two different regions of an image.

Another large class of techniques trains a classifier to detect images generated by either convolutional or generative adversarial networks. Examples include \cite{conv1}, which uses an expectation maximization algorithm (EM) to detect patterns of correlation among pixel neighborhoods. The correlation patterns are used to classify whether an image was generated by a convolutional network. Similarly, \cite{conv4} shows that images generated by a large suite convolution-based generators (including DeepFakes) can be detected by training a ResNet model on images from only a single generator, which suggests that these images exhibit common structure. Considering GANs instead of convolutional networks, in \cite{conv2} the residual of an autoencoder-based image reconstruction is deemed a ``GAN fingerprint'' and used to classify GAN-generated from authentic images. Along similar lines, \cite{conv3} uses a discrete Fourier transform to detect artifacts of the upsampling procedure used by many GAN-based image generators.

Some methods for classifying authentic from manipulated images use deep learning, but do not focusing on specific features of the DeepFake generation process. Examples include \cite{net1}, which emphasizes using a lower number of layers in its architecture, and \cite{net2}, which uses a capsule network as opposed to more traditional convolutional layers.

A final class of techniques uses features derived from knowledge that the image contains a human face. In \cite{human1}, manipulated videos are detected by noting the lack of natural blinking patterns. More recent DeepFakes circumvent this by including images where with a blinking subject in the training data. The authors of \cite{human2} detect manipulated images using inconsistencies in the eye and teeth regions, where lack of detail is common. In \cite{human3}, the authors consider the problem of determining whether a video contains a certain individual, such as a public figure. The individual's facial mannerisms are captured using a correlation matrix of facial actions, such as eyebrow or chin raising, across frames in a video, and this correlation matrix is used as features in an SVM to determine authentic from manipulated videos. In follow-up work \cite{farid}, the authors use a convolutional network to capture facial mannerisms, which avoids the labor-intensive process of hand-crafting the correlation-based features. Finally, \cite{human4} estimates blood volume changes that occur due to a rhythmic heartbeat, and uses this information to distinguish authentic from synthetic videos. The analytic we consider in this paper falls into this final class of techniques that leverage the fact that the image contains a human subject, because its method for distinguishing real from manipulated images utilizes the structure of the human face.




\section{Review of Analytic} \label{sec:review}

In this section we will review the analytic, which detects a DeepFake image or video frame using inconsistencies between the inner (swapped) region of a face and the outer (unaltered) region. The analytic computes two 3D head pose estimates, one using only the central region of the face and the other its entirety, and using various features derived from these head poses classifies an image as either manipulated or authentic. The authors of the analytic claim that, because only the inner portion of a DeepFake image is manipulated, inconsistencies can be detected this way, despite not being detectable to a human observer \cite[\S 1]{thepaper}.


In this section, we review the main steps of the analytic: detecting faces, locating facial landmarks, estimating 3D head poses, and training an SVM classifier on the resulting head poses. In each of these steps, we pay special attention to the techniques used and the assumptions required for valid application.

\subsection{Detecting Faces}

Given an image to be classified, the first step in the analytic is to detect faces in the image. To do so, the analytic uses a Histogram of Oriented Gradients (HOG) \cite{HOG} face detector as implemented in the \texttt{dlib} computer vision library \cite{dlib}. If no faces are detected in the image, which frequently occurs when faces are not oriented towards the camera, the analytic exits without making a prediction. 

\subsection{Locating Facial Landmarks}

Once a face has been detected, the analytic estimates the position of 68 facial landmarks, 51 of which are in the inner region of the face. Thirty of the interior facial landmarks are not used by the analytic, like those outlining the lips and eyes, because they are dynamic and do not indicate the position and orientation of the head relative to the camera. To locate the facial landmarks, the analytic applies a well-regarded technique for landmark estimation \cite{landmarks}, which uses an ensemble of gradient-boosted regression trees to find the landmarks' positions. Because of the important role it plays in the analytic, we will describe this landmark detection technique in more detail.

We focus our attention on the application of a pretrained landmark estimation model, since this reflects its usage in the analytic. The landmark detection technique refines, for a fixed number of iterations $T$, an initial estimate $\hat{S}_{0} \in \mathbb{R}^{68 \times 2}$ of the $(x,y)$ pixel coordinates for the 68 facial landmarks. At each iteration, $k$, the landmark positions are updated as
$$\hat{S}_{k+1} = \hat{S}_{k} + \gamma \, r_{k}(\hat{S}_{k}, I)$$
where
\begin{itemize}
\item $I \in \mathbb{R}^{m \times n}$ is a gray-scale representation of the image or region of interest, with dimensions $m \times n$.
\item $\left\{r_{k}: \mathbb{R}^{68 \times 2} \times \mathbb{R}^{m \times n} \to \mathbb{R}^{68 \times 2}\right\}_{k = 1}^{T}$ is a sequence of random forest regression functions.
\item $\gamma \in (0,1)$ is a \emph{shrinkage parameter}, commonly used in boosted forests to mitigate overtraining.
\end{itemize}
Each random forest $r_{k}$ acts on $\hat{S}_{k}$ and $I$ using only a finite number of points in $I$, the location of which are fixed relative to $\hat{S}_{k}$'s deviation from a set of mean facial landmarks. Specifically, split nodes in the decision trees of $r_{k}$ are based on the criteria
$$ I\left(A(\hat{S}_{k}) u\right)- I\left(A(\hat{S}_{k}) v\right) > \tau $$
where $\tau \in \mathbb{R}$ and $u, v \in \mathbb{R}^{2}$ are fixed at each split node, and $A: \mathbb{R}^{68 \times 2} \to \mathbb{R}^{2 \times 2}$ is a function mapping the current landmark estimates $\hat{S}_{k}$ to a $2 \times 2$ matrix. This matrix is a similarity transform that maps from a set of mean facial landmarks to $\hat{S}_{k}$.

We note a couple aspects of this landmark detection technique which are relevant to its application in the analytic. First, the training data for the ensemble are exclusively unmanipulated faces and their landmarks, so that its performance on DeepFakes remains to be validated. Second, the landmark refinements rely on an ensemble of decision trees, with split nodes depending on the current landmark estimate and a pair of grey-scale pixels $u$ and $v$. These points are optimally selected from a random sample during training, with the important consequence that $u$ and $v$ need not respect the face's local geometry. For example, $u$ and $v$ both on a subject's chin have the potential to affect landmark refinements around the nose. This does not respect the analytic's assumption that the estimated position of outer landmarks will be minimally affected by the manipulation of the inner facial region, a point which we revisit in section \ref{sec:assumps}.

\subsection{Estimating 3D Head Poses} \label{sec:headpose_right}

After the 68 facial landmarks have been estimated, the analytic uses a pinhole camera model, combined with some assumptions about the camera's focal length and 3D geometry of the subject's head, to estimate the head's orientation and position relative to the camera. Two of these estimates are taken, one using all the facial landmarks and the other using those in the inner region of the face. In this section we summarize this pose estimation procedure.

In order to estimate the head pose of a subject using a set of landmarks, the analytic considers the error between the set of 2D facial landmarks detected in the image and a set of 3D facial landmarks from an ``average'' model of the human face \cite{openface}, which are projected into the image using the pose estimate and the geometry of the pinhole camera model. The subject's head pose is estimated by minimizing the error between the projected 3D landmarks and the detected 2D ones over all possible head poses. 

Let $\mathcal{L}$ index a collection of landmarks. For our purposes, $\mathcal{L}$ will either index all of the landmarks or those in the inner region of the face. Denote by $\left\{(x_{i}, y_{i}) \right\}_{i \in \mathcal{L}}$ the landmarks detected in the image and $\left\{(U_{i}, V_{i}, W_{i})\right\}_{i \in \mathcal{L}}$ the corresponding landmarks from an average 3D model of the human face. The 3D facial landmarks are given in a basis with the face's center as the origin and eyes looking in the positive $z$ direction. The analytic then estimates the head pose of the subject relative to the camera by finding the optimal rotation matrix $R$ and translation vector $t$ such that the $\left\{(U_{i}, V_{i}, W_{i})\right\}_{i \in \mathcal{L}}$ coordinates align with the landmarks detected in the image. That is, the analytic minimizes
\begin{equation}
\label{Opt}
\footnotesize
\sum_{i\in \mathcal{L}}\left\|s\left[\begin{array}{c}
x_{i} \\
y_{i} \\
1
\end{array}\right]-\left[\begin{array}{ccc}
f_{x} & 0 & c_{x} \\
0 & f_{y} & c_{y} \\
0 & 0 & 1
\end{array}\right]\left(R\left[\begin{array}{c}
U_{i} \\
V_{i} \\
W_{i}
\end{array}\right]+t\right)\right\|^{2}
\end{equation}
over rotation matrices $R \in \mathbb{R}^{3 \times 3}$, translation vectors $t \in \mathbb{R}^{3}$, and projective scaling $s \in \mathbb{R}$. In this problem, the upper triangular matrix is the \emph{camera matrix} which projects a 3D object into pixel coordinates. The $f_{x}$ and $f_{y}$ values denote the camera's focal length multiplied by the pixel density in the $x$ and $y$ dimensions, respectively. Together, $(c_{x}, c_{y})$ give the principal point of the image in pixel units. In summary, the problem in \eqref{Opt} minimizes the projection error between the image landmarks and the 3D coordinates as projected into the image. For more information on the geometry of the pinhole camera model we refer the reader to \cite{pinhole}.

The head pose estimation problem in \eqref{Opt} relies on a number of important assumptions. First, one might expect that detecting inconsistencies in inner facial landmarks requires estimating the position of these landmarks to a reasonably high accuracy, and using an average 3D model of the human face as a substitute for the subject's face may undermine that accuracy. Second, the camera matrix must be estimated because these physical properties of the camera cannot be derived from a single image. When the image considered is of resolution $m \times n$, the analytic approximates the principal point with the center of the image, $c_{x} = m/2$, $c_{y} = n/2$, and the focal length terms with $f_{x} = f_{y} = m$. Third, computing minima in \eqref{Opt} is nontrivial because it requires optimizing over the nonconvex set of 3D rotation matrices, and the analytic assumes the head poses it computes are accurate. The analytic circumvents this concern by using a well-regarded method to solve this problem, as implemented in \texttt{opencv} \cite{opencv}. This optimization algorithm initializes values of $R$ and $t$ by computing a Direct Linear Transform, in which the rotation matrix constraint on $R$ is relaxed. The relaxed solution is projected onto the set of rotation matrices, after which an iterative Levenberg-Marquardt algorithm is applied to further reduce the objective function \cite{levenberg}.

\subsection{Training the Classifier} 

The final step in the analytic uses features derived from the head pose estimates--an $R$ matrix and $t$ vector for both inner and all landmark pose estimates--to classify an image as DeepFake or authentic. The analytic uses a support vector machine with radial basis function kernel to perform the classification. In their original paper, the authors test various features constructed from the $R$ matrices and $t$ vectors of the inner and all landmark estimates, ultimately concluding that the flattened difference between the $R$ matrices for each head pose estimate, with the difference in $t$ vectors appended, yields superior classification results. The $\gamma$ parameter in the radial basis function kernel $K(x,y) = \exp(-\gamma\|x-y\|^2)$ is fixed at $1/(\text{number of features})$, and the SVM model is trained to output class probabilities using Platt scaling \cite{platt}.

\section{Problematic Assumptions} \label{sec:assumps}

In this section we introduce a number of methodological issues related to the analytic. We do so by formulating and then refuting various, often implicit, assumptions justifying its use.

\subsection{Landmark Estimates}

\begin{figure}[h]
\begin{center}
\includegraphics[width=.2495\textwidth]{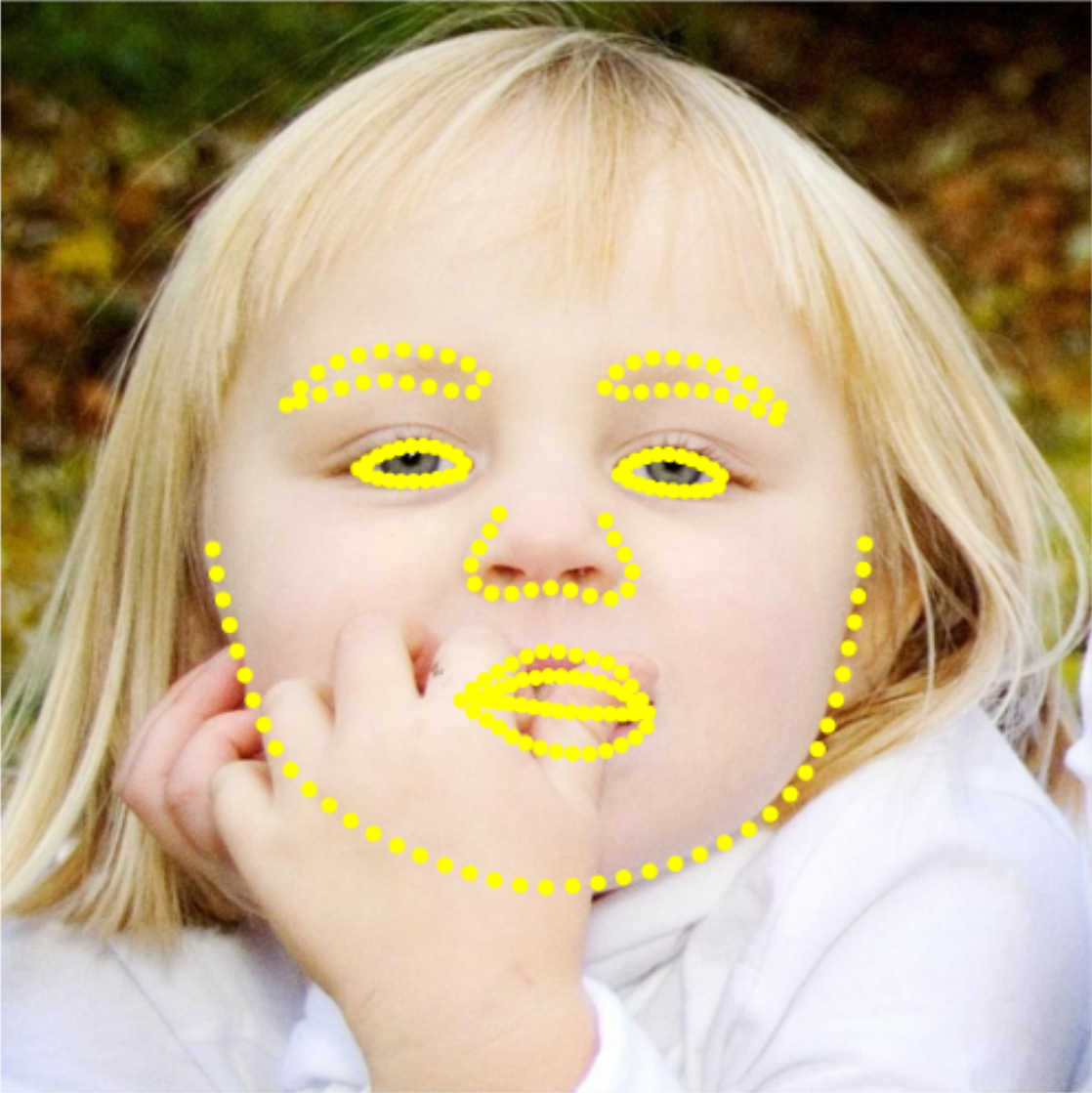}%
\includegraphics[width=.25\textwidth]{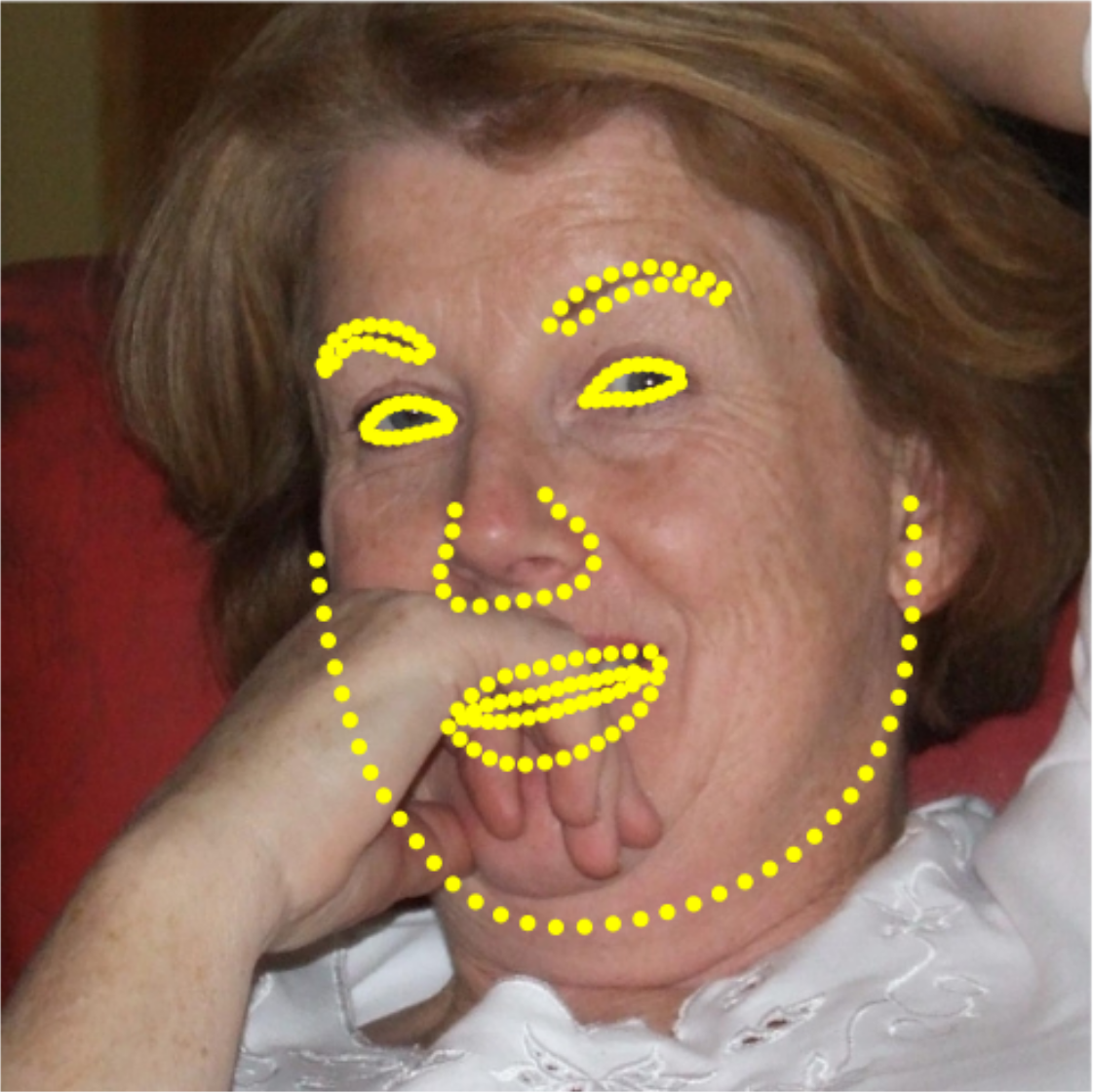}
\end{center}
\caption{The analytic's landmark estimates are designed to overcome locally inconsistent regions, such as facial obstructions. Source: \cite{landmarks}.}
\label{fig:landmarkpaper}
\end{figure}

The motivating principle behind the analytic--that DeepFakes introduce inconsistency between the inner and outer regions of the face--relies on the following assumption.
\begin{assumption} \label{A1} Landmark estimates in the inner region of the face are affected by manipulations of that region, while landmark estimates outside of the inner region are minimally affected.
\end{assumption}
This assumption makes the restriction to inner landmarks meaningful, because it implies that only the inner landmark estimates will change in a response to a splice in the inner facial region. Though this assumption seems intuitive, the landmark estimation technique the analytic uses is not designed for use on manipulated images. In the context of authentic images, it is \emph{desirable} for a landmark estimation method to predict realistic landmark estimates despite inconsistencies in certain regions of the face. These inconsistencies often occur due to obstructions, such as sunglasses or hair styles. Figure \ref{fig:landmarkpaper} demonstrates this point, where the estimation method from the analytic overcomes facial obstructions to produce landmark estimates which appear to be consistent with unobstructed regions of the face. This suggests that the assumed insensitivity of outer landmark estimates to manipulations in the inner face is in conflict with the design goals of landmark estimation in general.

\begin{figure}[h]
\begin{center}
\includegraphics[width=.25\textwidth]{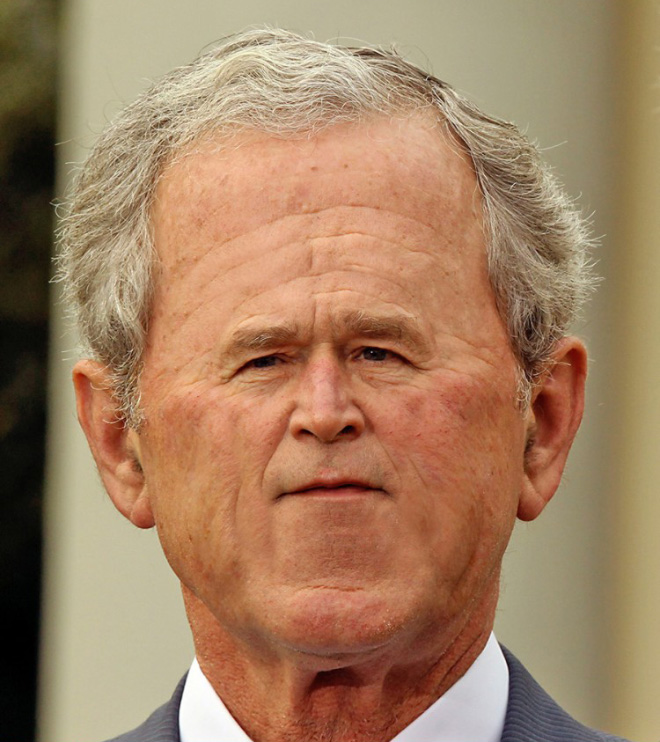}%
\includegraphics[width=.25\textwidth]{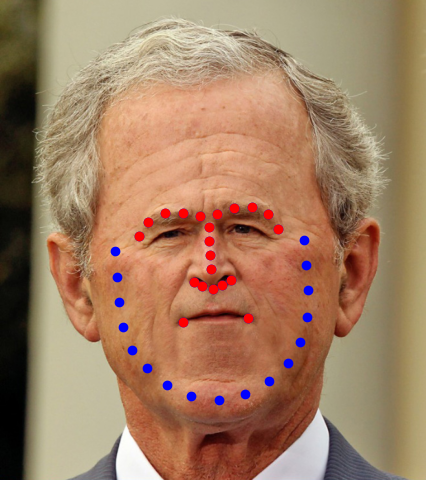}\\
\includegraphics[width=.25\textwidth]{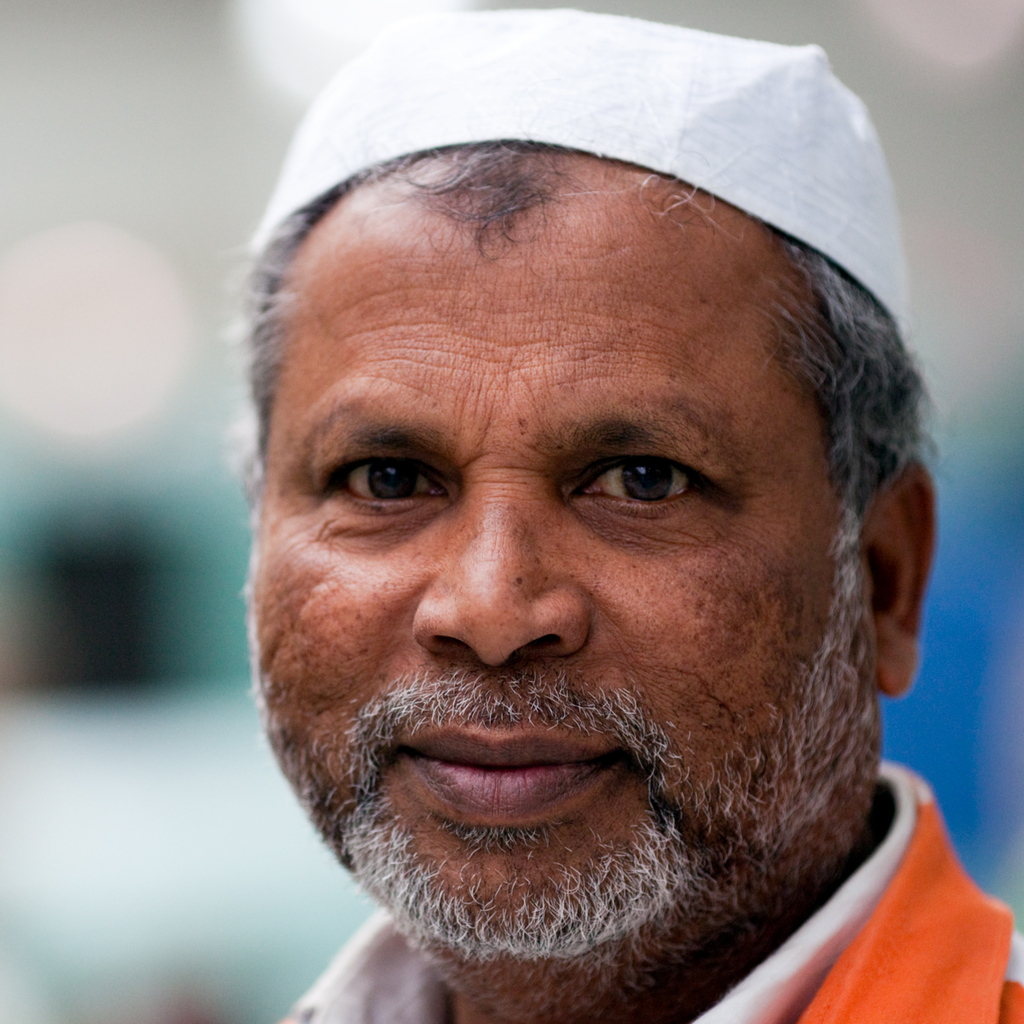}%
\includegraphics[width=.25\textwidth]{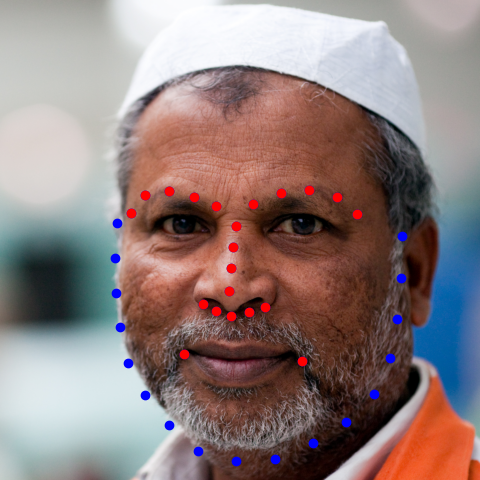}
\end{center}
\caption{(Top) Manipulated inner face (red) leads to outer landmarks (blue) which are incorrectly located. (Bottom) Authentic image and landmarks. Sources: \cite{smallbush} and \cite{FFHQ}.}
\label{fig:tinyface}
\end{figure}

To test the validity of Assumption \ref{A1}, we estimate the facial landmarks in an image where the inner region of the face is obviously and dramatically inconsistent with the outer region. Figure \ref{fig:tinyface} gives a manipulated image, with an authentic image for comparison. The inner region of the manipulated image is shrunk beyond reasonable proportions, while the outer region of the face is left unaltered. The outer landmark estimates do not extend to the outer region of the face as they do in the authentic image, because the predictive model produces landmark estimates similar to the authentic landmarks used to train the model. These results undermine Assumption \ref{A1}, and suggest that landmark detectors designed for authentic images may not reliably detect inconsistencies in manipulated images.

\subsection{Head Pose Optimization} \label{sec:headpose}

For a given collection of 2D landmarks detected in the image and 3D landmarks from a model of the typical human face, minimizing \eqref{Opt} over $R$, $t$, and $s$ yields an estimate of the head pose. The head's orientation is given by rotation matrix $R$, and the head's location by the translation vector $t$. This is a special case of the classical \emph{Perspective-$n$-Point} (PnP) problem, which estimates an object's pose using 3D points, their projections into the image, and a known camera matrix. Because of its important role in various computer vision applications, many algorithms have been introduced to solve the PnP problem. By relying on the accuracy of an existing and well-reputed solver for the PnP problem \cite{solver}, the analytic assumes the following.
\begin{assumption}\label{A2}
The rotation matrix $R$ and translation vector $t$ obtained from minimizing \eqref{Opt} are accurate estimates of the head's true orientation and position relative to the camera.
\end{assumption}
We show next that general purpose PnP solvers should not be applied to human faces without special consideration for their symmetry. These observations demonstrate that the analytic's Assumption \ref{A2} does not hold.

Consider, for some choice of $R$ and $t$, the transformed 3D landmarks and their projections into the image.
$$\left[ \begin{array}{c} X_{i} \\ Y_{i} \\ Z_{i} \end{array} \right] := R \left[ \begin{array}{c} U_{i} \\ V_{i} \\ W_{i} \end{array} \right] + t$$
$$\left[\begin{array}{c} \hat{x}_{i} \\ \hat{y}_{i} \end{array} \right] = \left[ \begin{array}{c} \frac{X_{i}}{Z_{i}} f_{x} + c_{x}\\ \frac{Y_{i}}{Z_{i}} f_{y} + c_{y} \end{array} \right]$$
Note that $\hat{x}_{i}$ and $\hat{y}_{i}$ are invariant with respect to the transformation $(X_{i}, Y_{i}, Z_{i}) \to (-X_{i}, -Y_{i}, -Z_{i})$. By increasing $R$'s yaw rotation by $\pi$ and adjusting the $z$-component of $t$, we can approximately perform this transformation up to a slight deviation that can be attributed to the 3D landmarks' deviation from planarity. In this way we can construct, for each choice of $R$ and $t$, an estimate with nearly identical projection error.

\begin{figure}[h]
\begin{center}
\includegraphics[width=.43\textwidth]{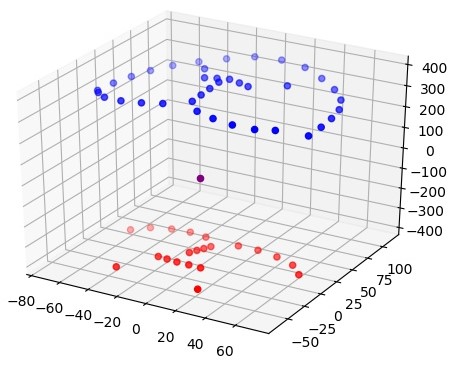}
\end{center}
\caption{Oriented 3D landmarks for both inner (red) and all (blue) landmark estimates, with the pinhole camera aperture in purple at the origin. In this example, the conflicting head pose estimates illustrate the two local minima of \eqref{Opt}.}
\label{fig:flipped}
\end{figure}

This observation has important consequences for the validity of Assumption \ref{A2}, because it shows that the optimization problem \eqref{Opt} has at least two local minima with similar squared projection error. Figure \ref{fig:flipped} shows the transformed 3D landmarks in their estimated poses, for both the inner and all landmark estimates. These two pose estimates demonstrate the two local minima of \eqref{Opt}. Even though the pose estimate using all landmarks suggests that the landmarks are visible through the back of the subject's head, this solution is consistent with the provided landmarks and the geometry of the pinhole camera model.

Further experiments demonstrate that the analytic frequently uses head poses which suggest the image was captured through the back of the subject's head. We refer to these poses as \emph{flipped}, and refer to a pair of head pose estimates with one correct and one flipped as  \emph{conflicting} head pose estimates. Table \ref{tab:flips} demonstrates that, for the University of Albany Deep Fake Video (UADFV) data used by its authors to demonstrate the analytic's performance, $96\%$ of video frames in the training set had at least one flipped pose estimate and  $33\%$ had conflicting pose estimates. The high proportion of flipped poses used in the analytic invalidates Assumption \ref{A2}.

\begin{table}[h]
\begin{center}
\begin{tabular}{c  c | c | c |} \cline{3-4}
& & \multicolumn{2}{c|}{inner landmark} \\ \cline{3-4}
& & flipped & correct \\ \cline{3-4}
\hline \multicolumn{1}{|c|}{\multirow{2}{4em}{all landmark}} & flipped & .63 & .28\\ \cline{2-4}
\multicolumn{1}{|c|}{} & correct & .05 & .04 \\ \hline
\end{tabular}
\caption{The proportion of head pose estimates in the UADFV training set which exhibit a flipped pose.}
\label{tab:flips}
\end{center}
\end{table}

Detecting a flipped head pose estimate is straightforward by examining the sign of $Z_{i}$, the $z$ component of the transformed 3D landmarks. Alternatively, because the 3D landmarks have $W_{i} \approx 0$, the sign of the translation vector's $z$ component will suffice. Figure \ref{fig:flipped} illustrates this point; the correct head pose has $Z_{i} < 0$, whereas the flipped pose has $Z_{i} > 0$ for all landmarks. Furthermore, we can modify the head pose estimation algorithm from \ref{sec:headpose_right} to avoid the local minimum of a flipped head poses by checking for $Z_{i} > 0$ and correcting the yaw and translation vector. This step can be naturally incorporated immediately after projecting the DLT solution onto the set of rotation matrices, or after some amount of Levenberg-Marquardt iterations have been applied. We opt for the latter, and refer to this as the corrected version of the analytic in the remainder.

One implication of the analytic's use of flipped head pose estimates is that any trends which correlate with conflicting pose estimates will be easy to identify. For the UADFV data set, manipulated frames are four times more likely to have conflicting pose estimates than authentic frames. Since the features used in the SVM are the flattened differences of the rotation matrices and translation vectors, conflicting pose estimates are easy for the classifier to identify.

\subsection{Overtraining and Algorithmic Bias} \label{subsec:algobias}

\begin{figure}[h]
\begin{center}
\includegraphics[width=.25\textwidth]{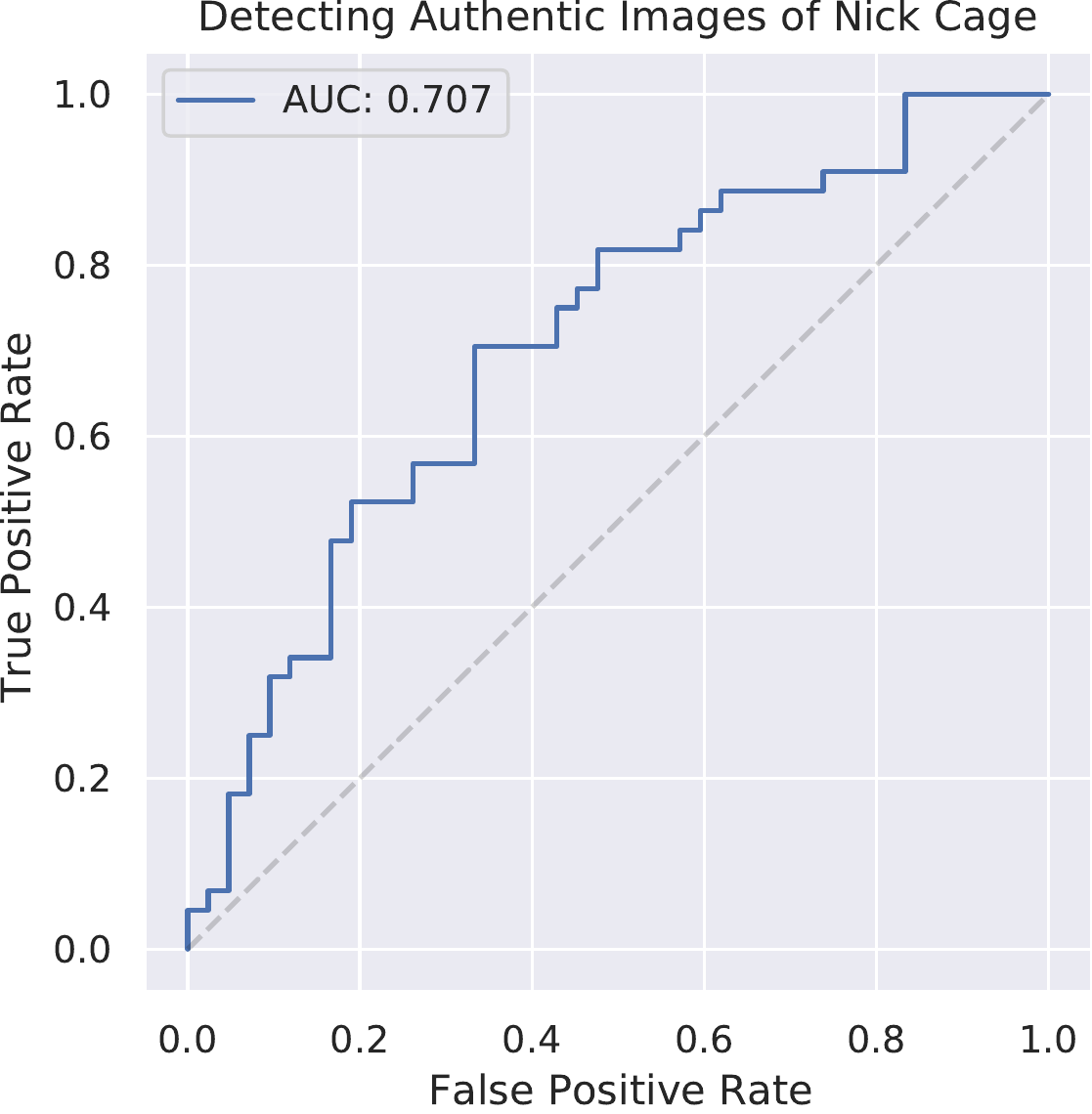}%
\includegraphics[width=.25\textwidth]{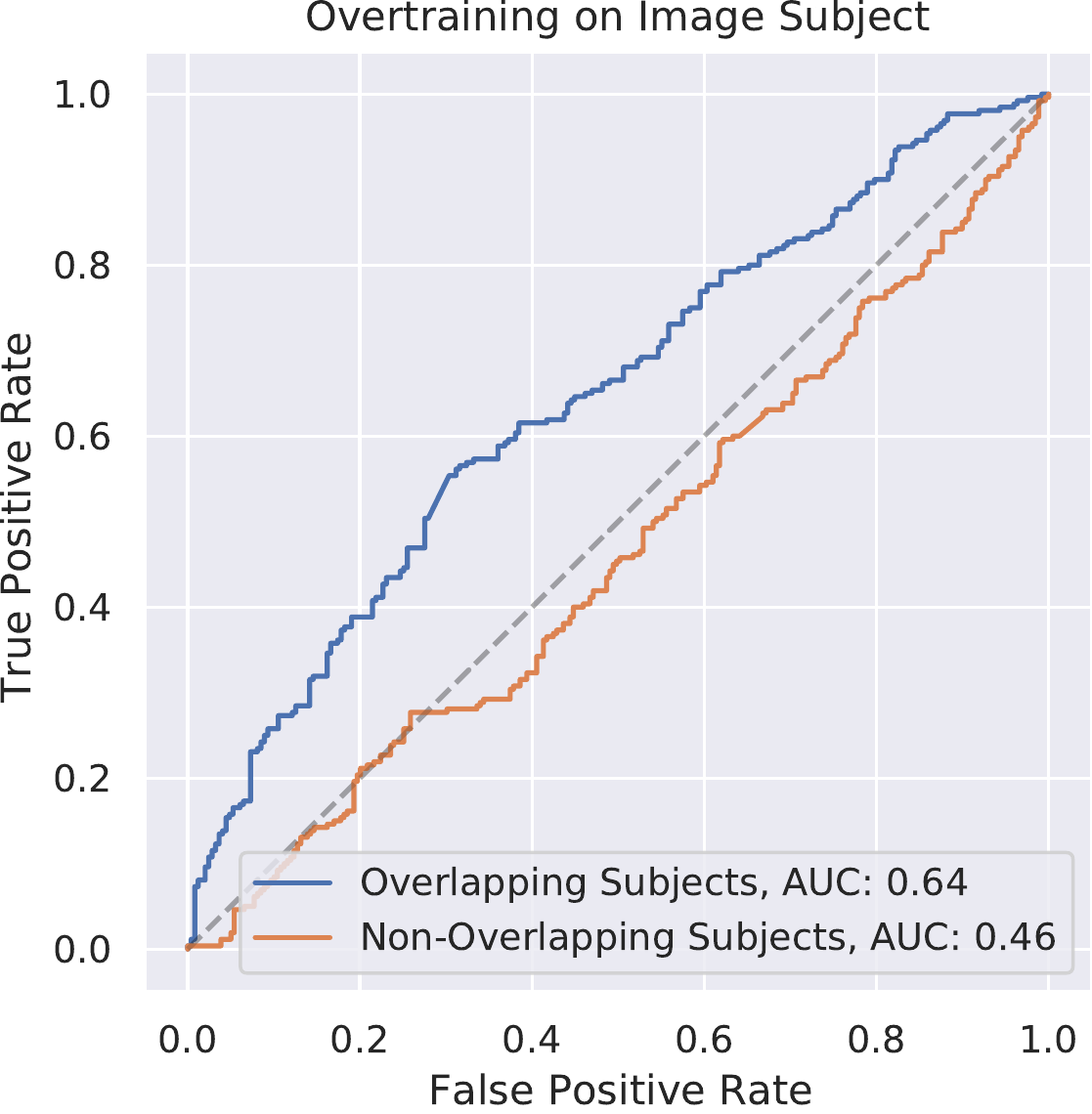}
\end{center}
\caption{(Left) The authors' pretrained model classifies authentic images of Nicolas Cage as manipulated because he is depicted in all of the manipulated images of the UADFV training set. (Right) Performance of two models on images from DFDC indicate that the analytic overtrains based on the depicted individual. In the Overlapping Subjects model, subjects appear in either manipulated or authentic images across both the training and testing sets. In the Non-Overlapping Subjects model, the training and testing sets feature disjoint subjects.}
\label{fig:Cage_and_DFDC}
\end{figure}

Though certain methods for DeepFake detection are concerned with deciding whether an image or video contains a person of interest, this analytic does not make such a restriction. As such, the analytic should generalize to individuals not used in the training set.
\begin{assumption}\label{A4}
The identity of the image's subject has no bearing on its predicted label.
\end{assumption}
One aspect of the analytic that invites further scrutiny is the data used to demonstrate its performance. The UADFV data set used by the analytic's authors swaps a single individual's face--that of actor Nicolas Cage--into all of the manipulated images. A simple test shows that this results in a model trained to detect images of Nicolas Cage and not DeepFakes. Figure \ref{fig:Cage_and_DFDC} shows an ROC curve for a test set of 111 images, all of which are authentic, where 56 of the images contain Nicolas Cage and 55 contain other celebrities. Applying the authors' pretrained model to these images shows that the analytic reliably identifies authentic images of Nicolas Cage as manipulated.

A follow-up experiment confirms the analytic's tendency to overtrain based on identity of the image's subject. Figure \ref{fig:Cage_and_DFDC} contains ROC curves for two train/test splits of video frames from the DeepFake Detection Challenge (DFDC) data set \cite{DFDC}. In the first split, each subject appears in either manipulated or authentic images, but not both, across the training and testing sets. The second split is like the first, except that the sets are additionally restricted so that testing and training images have no subjects in common. We perform the experiment using a modification of the analytic that includes the head pose correction detailed in Section \ref{sec:headpose}, but note that the uncorrected results are similar and culminate in the same conclusion. The results indicate that the model AUC increases by $.18$ when the context in which a subject appears (manipulated or authentic images) persists across the training and testing sets. Moreover, when the subjects in the testing set are disjoint from the training image, the model's performance is roughly equivalent to choosing the predicted label uniformly at random. 

These results demonstrate that Assumption \ref{A4} is invalid, because the analytic demonstrates the tendency to overtrain based on the identity of the subjects in an image. This is especially concerning because of the analytic's potential to classify images based on shared facial structure as those in the training set, which may be due to attributes like age, race, or gender of the images' subjects.

\subsection{DeepFakes Exhibit Inconsistent Head Orientation}

In order for the analytic to distinguish DeepFakes from authentic images, the distribution of head poses should differ between real and manipulated images.

\begin{assumption}\label{A3}
The distribution of head poses differs between DeepFakes and authentic images.
\end{assumption}

\begin{figure}
\includegraphics[width=.25\textwidth]{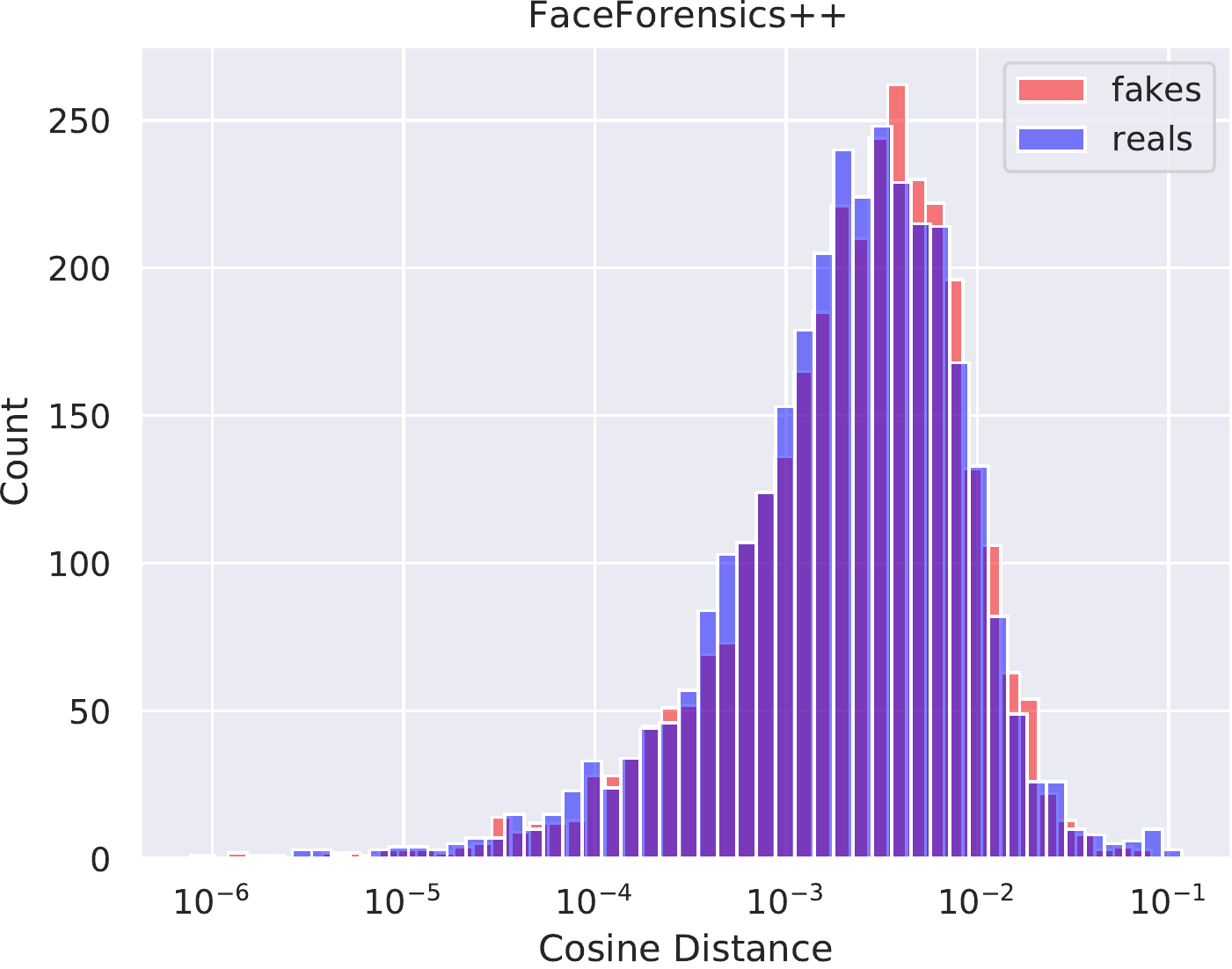}%
\includegraphics[width=.25\textwidth]{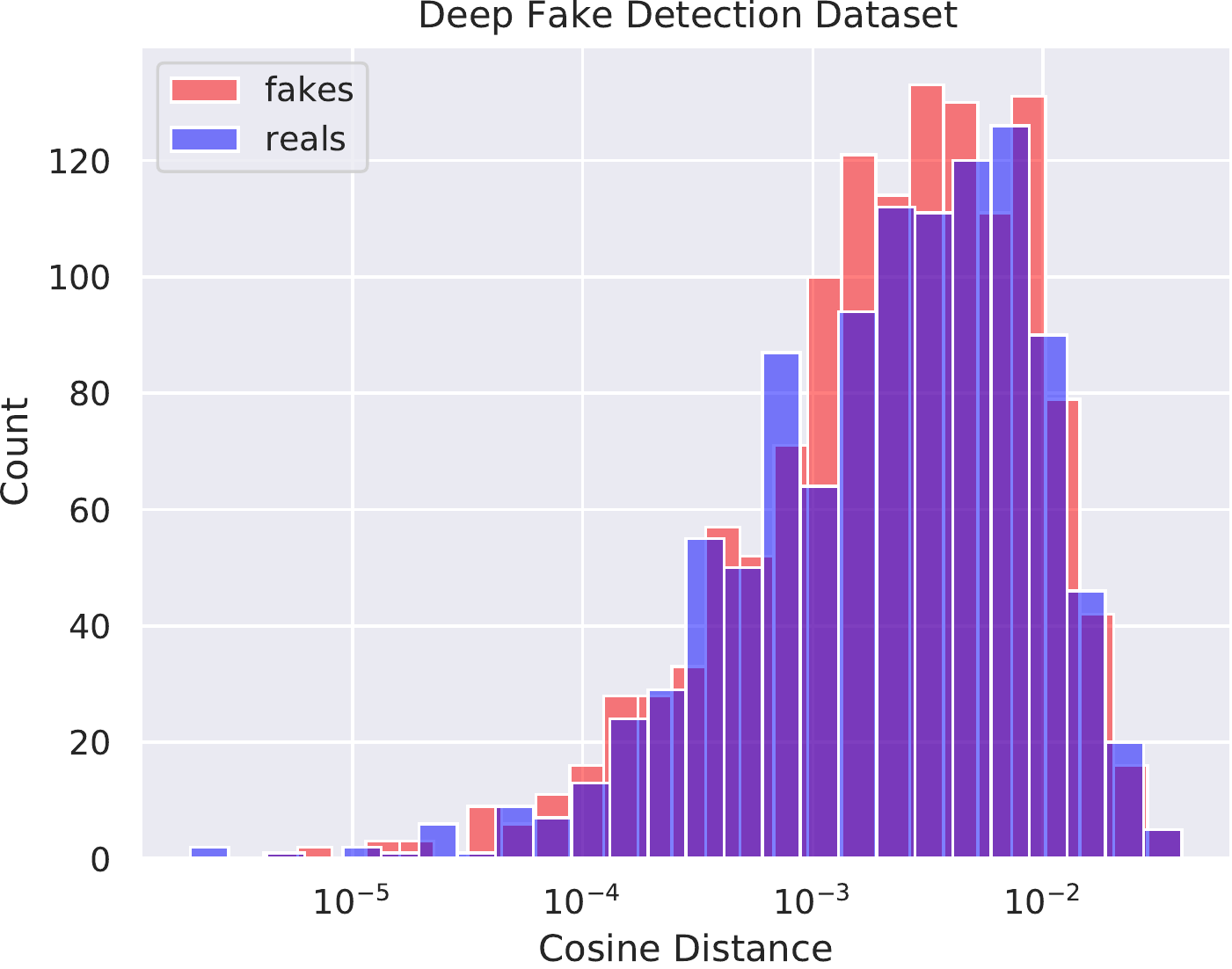}\\

\vspace{.05cm}
\includegraphics[width=.25\textwidth]{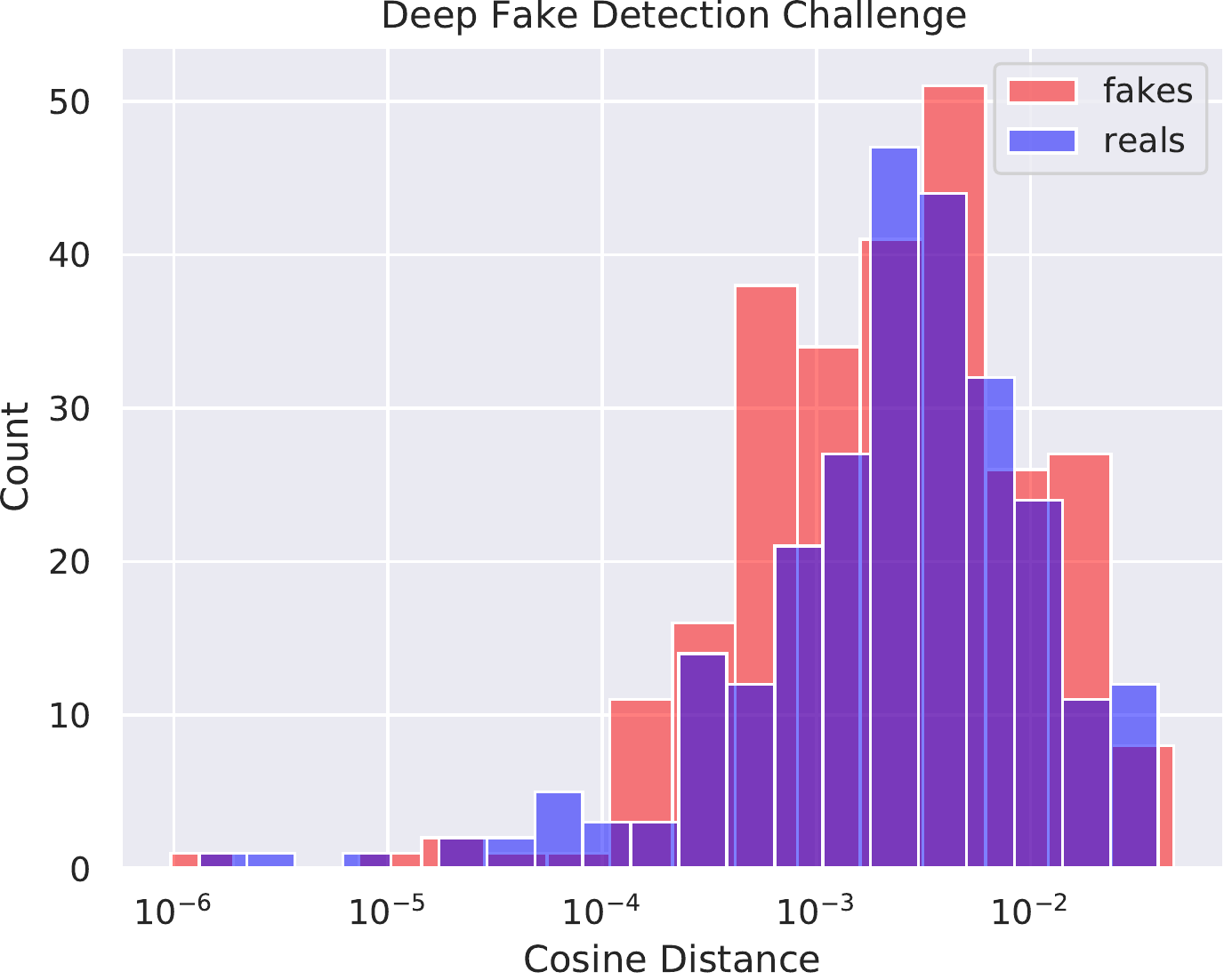}%
\includegraphics[width=.25\textwidth]{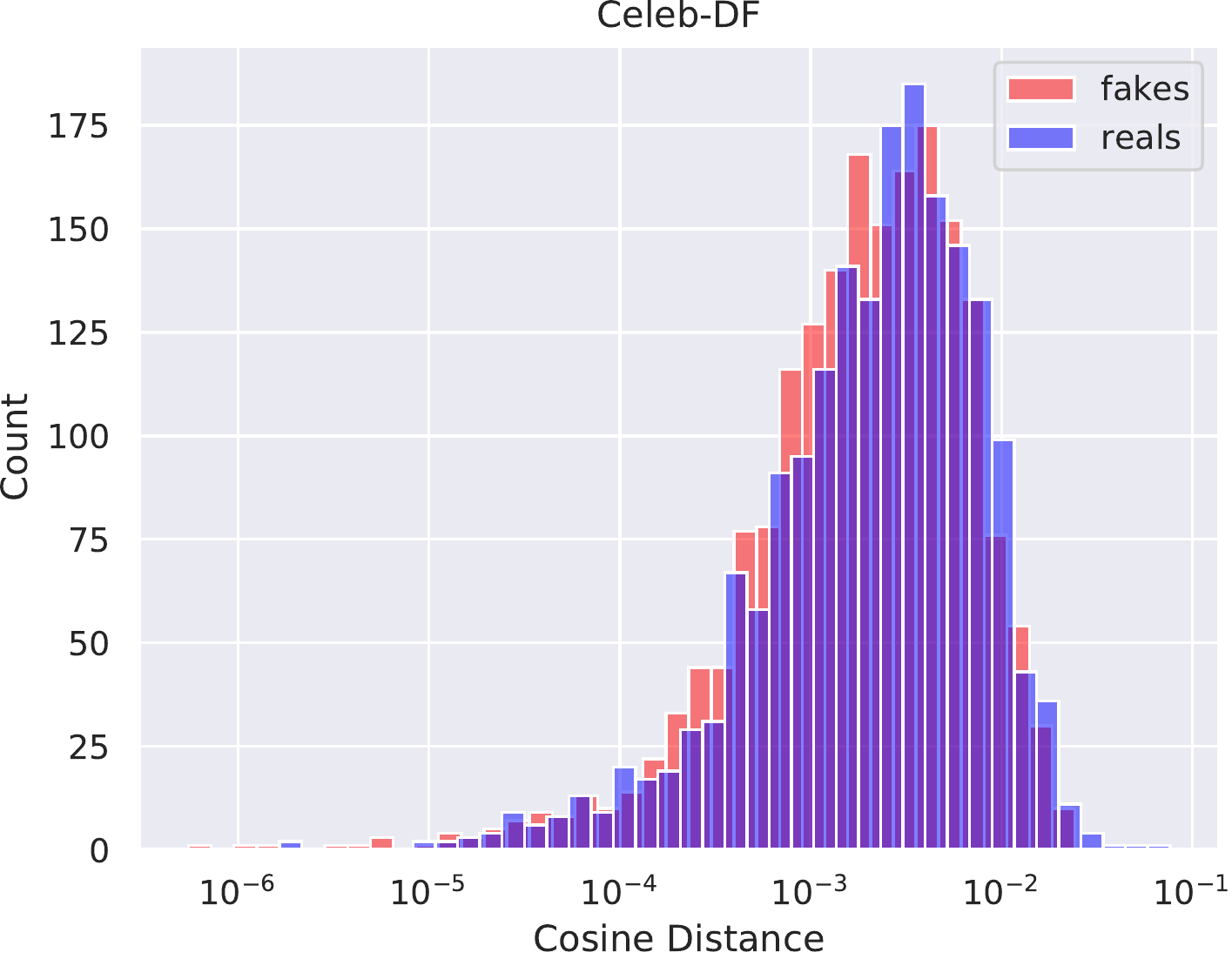}
\caption{Histograms of cosine distances between ``inner'' and ``all'' head orientation vectors. Contrary to the analytic's Assumption \ref{A3}, this feature does not separate DeepFakes from authentic images.}
\label{fig:cosdists}
\end{figure}

In the paper introducing the analytic, the authors claim that, with $R_{c}$ and $R_{a}$ denoting the estimated rotation matrices for the central landmark and all landmark estimates, respectively, the cosine distances of the head orientation vector can be used to separate DeepFakes from authentic images. That is, for $w = [0, 0, 1]^{\top}$,
\begin{equation}\label{eq:cosdist}
 1 - \frac{\langle R_{a}^{\top} w, R_{c}^{\top}w \rangle}{\left\|R_{a}^{\top} w\right\| \left\|R_{c}^{\top} w\right\|}
\end{equation}
separates DeepFakes and authentic images. We find this claim, and the more general Assumption \ref{A3}, to be unreproducible across all the data sets considered, and regardless of whether the uncorrected or corrected analytic is used. Additionally, due to the occurrence of flipped poses, we find that the cosine distance metric is not as effective as the authors intend since the metric does not detect the difference in yaw rotation between conflicting pose estimates. Figure \ref{fig:cosdists} shows the histograms of the cosine distances between head orientation vectors for various DeepFake data sets introduced in section \ref{sec:reprod}.

Additional investigations yield no evidence of separation between the DeepFake and authentic image classes for any of the twelve underlying estimated parameters (roll, pitch, yaw, and a 3D translation vector for both head pose estimates) for any of the data sets in table \ref{tab:splits}, when considering either the original head pose estimation procedure or our corrected version. Moreover, table \ref{tab:more_flips} shows that the correlation of flipped head poses with DeepFakes in UADFV does not hold in general. The exact cause of this correlation in UADFV is unknown, but we conjecture that certain faces (i.e. that of Nicolas Cage) are more susceptible to the local optimum in \eqref{Opt} than others.
\begin{table}
\begin{center}
\begin{tabular}{r|cc}
& $P(\text{fake}|\text{conflicting})$ & $P(\text{fake}|\text{non-conflicting})$\\ \hline
UADFV & $.78$ & $.38$\\
DARPA & $.64$ & $.39$\\
FF++ & $.47$ & $.51$\\
DFDD & $.55$ & $.54$\\
DFDC & $.49$ & $.50$\\
Celeb-DF & $.49$ & $.51$
\end{tabular}
\end{center}
\caption{The empirical probability of an image being manipulated given that the pose estimates are conflicting. In this table, ``conflicting'' denotes the case that exactly one of the pose estimates is flipped, which is easily distinguishable using the translation vector feature of the SVM. The relationship between conflicting pose estimates and DeepFakes in UADFV and DARPA-GAN does not generalize.}
\label{tab:more_flips}
\end{table}
These observations suggest that Assumption \ref{A3} may not hold, and brings the utility of features related to estimated head poses into question.

\section{Reproducibility Study} \label{sec:reprod}

This section contains a large scale reproducibility study aimed at measuring the analytic's ability to generalize to new data sets. The various problematic assumptions discussed in section \ref{sec:assumps} suggest that, despite its celebrated status, many of the analytic's motivating principles are flawed. This section continues this investigation by comparing its predictive performance on different collections of DeepFakes.

First, we note that we are able to reproduce the authors' results on the UADFV data set used in the original paper, using both the authors' pretrained model and one which we trained ourselves using the authors' training split. Both models have a near-perfect AUC of approximately $.98$. Correcting for the flipped head poses from section \ref{sec:headpose} reduces the performance of these models to $.88$, as the model cannot use UADFV's correlation between flipped head poses and fake images observed in table \ref{tab:more_flips}. However, the problematic nature of the UADFV data set discussed in the previous sections prompts us to look at other data sets to test the analytic's utility as a DeepFake detector.

\begin{figure}
\begin{center}
\includegraphics[width=.2\textwidth]{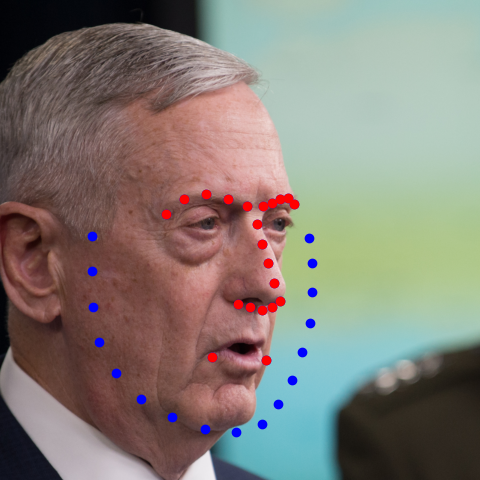}%
\includegraphics[width=.2\textwidth]{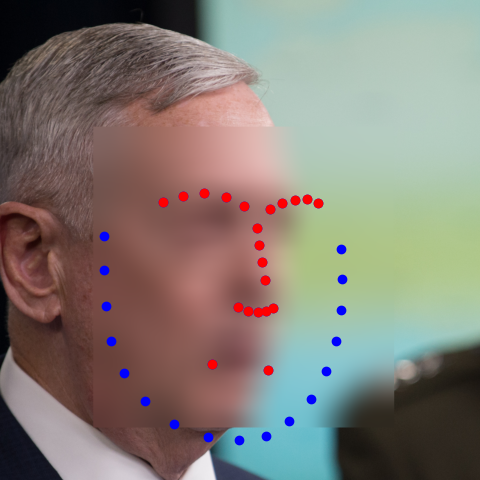}
\end{center}
\caption{Blurry faces, commonly found in low-quality DeepFakes, make landmark estimation unreliable. Source: \cite{FFHQ}.}
\label{fig:blur}
\end{figure}

In addition to  UADFV, the analytics' authors also consider another data set, a subset of images from the DARPA GAN challenge. However, the authors acknowledge that the faces shown in these DeepFakes are often extremely blurry, which makes landmark estimation unreliable. Figure \ref{fig:blur} illustrates this point, with the landmark estimates degrading as the blur increases in the facial region. 
In addition, table \ref{tab:more_flips} demonstrates that the DARPA data exhibit a similar correlation between conflicting head poses and fake images as UADFV, a trend which does not generalize. We also find that the DARPA data set has many near-duplicate images, suggestive of consecutive video frames, which in combination with its lower sample size ($<500$ images) greatly undermines its quality. Though we are able to reproduce the authors' results on the DARPA GAN data set, we do not consider it further because of these limitations.

We compare the analytic's performance on four modern DeepFake data sets: FaceForensics++ (FF++) \cite{FF++}, Deep Fake Detection Dataset (DFDD) \cite{DFDD}, Deep Fake Detection Challenge (DFDC) \cite{DFDC}, and Celeb-DF \cite{Celeb-DF}. Table \ref{tab:splits} gives the number of randomly selected frames in our training and testing sets for each data set. Because of the analytic's tendency to overtrain based on the identity of the image's subject, whenever the data allows we construct a random split of the data such that subjects are disjoint across the training and testing sets. The only data set which does not permit such a split is DFDD, which features 28 actors in manipulated and authentic videos, acting in a number of scenarios. For this data set, we randomly partitioned into training and testing set by scenario, while also ensuring that each actor occurs as both an authentic and manipulated subject in at least one video from each of the training and testing sets. In our experience, the analytic is sensitive to imbalanced data, so each set contains an equal number of authentic and manipulated images. 

\begin{table}
\begin{center}
\begin{tabular}{r|cc}
&  Training Frames  & Testing Frames \\ \hline
FF++ & 6392 & 1608 \\
DFDD & 2632 & 372 \\
DFDC & 600 & 600 \\
Celeb-DF & 3766 & 944 \\
\end{tabular}
\end{center}
\caption{Details of Train/Test Splits}
\label{tab:splits}
\end{table}

\begin{table}
\begin{center}
\begin{tabular}{r|ccc}
&  Provided  & Uncorrected & Corrected \\ \hline
FF++ & .52 & .60 & .63 \\
DFDD &  .46 & .47 & .28 \\
DFDC & .54 & .45 & .45 \\
Celeb-DF & .50 & .58 &59 \\
\end{tabular}
\end{center}
\caption{AUC scores for various models. The ``Provided'' model is pretrained on UADFV and provided by the authors \cite{bitbucket}. The ``Uncorrected'' model is trained on the training frames from table \ref{tab:splits}, without the head pose correction in section \ref{sec:headpose}. The ``Corrected'' model is also trained on the frames in table \ref{tab:splits}, but uses the corrected head pose model described in section \ref{sec:headpose}.}
\label{tab:AUCs}
\vspace{-.25cm}
\end{table}

Table \ref{tab:AUCs} gives our results, where we see that the analytic's performance is approximately what would be expected by assigning a predicted label to each image uniformly at random. The catastrophic performance of the corrected model on the DFDD data suggests that the analytic is again overtraining on the image's subject, despite our attempts to mitigate it. The corrected model appears to be especially sensitive to overtraining because it estimates the head pose more accurately than the uncorrected model.

\section{Conclusion} \label{sec:conclude}

In this work we show that a celebrated method for detecting DeepFakes is not reproducible. This study extends well beyond a set of numerical experiments, because we introduce a number of methodological flaws associated with using head pose estimates to detect manipulated images. We demonstrate that the features derived from this procedure contain information about the identity of the subjects contained in the training set, which raises privacy concerns for resulting classifiers. We also show that the approximate planarity of the facial landmarks considered leads to a strong local optimum in the 3D pose estimation problem, which skewed the originally presented results. 

We hope that our contributions will allow the performance of new DeepFake detectors to be properly interpreted, in addition to encouraging the development of these detectors in more productive directions.

{\small
\bibliographystyle{ieee_fullname}
\bibliography{bib}
}

\end{document}